\documentclass[12pt,a4paper]{article}

\usepackage{libertine}

\usepackage[T1]{fontenc}
\usepackage[utf8]{inputenc}
\usepackage{array}
\usepackage{comment}
\usepackage[table]{xcolor}
\usepackage{amsmath}

\usepackage[text={6.5in, 9in}, centering]{geometry}

\usepackage[printwatermark]{xwatermark}
\usepackage{wrapfig,lipsum,booktabs}

\usepackage{tabularx}
\usepackage{epstopdf}
\usepackage{subfigure}
\usepackage{graphicx}
\usepackage{etoolbox,hyperref}

\usepackage{authblk}
\usepackage[]{threeparttable}

\usepackage{array}
\newcolumntype{L}[1]{>{\raggedright\let\newline\\\arraybackslash\hspace{0pt}}m{#1}}
\newcolumntype{C}[1]{>{\centering\let\newline\\\arraybackslash}m{#1}}
\newcolumntype{R}[1]{>{\raggedleft\let\newline\\\arraybackslash\hspace{0pt}}m{#1}}

\usepackage{multicol}

\usepackage{booktabs} 

\begin{document}

\title{Deep Convolutional Neural Networks for Raman Spectrum Recognition : A Unified Solution}
\author[1]{Jinchao Liu}
\author[2]{Margarita Osadchy}
\author[3]{Lorna Ashton}
\author[4]{Michael Foster}
\author[1,5]{\\Christopher J. Solomon}
\author[1,5]{Stuart J. Gibson$^{\star}$}
\affil[1]{VisionMetric Ltd, Canterbury, Kent, UK}
\affil[2]{Department of Computer Science, University of Haifa, Mount Carmel, Haifa, Israel}
\affil[3]{Department of Chemistry, Lancaster University, Bailrigg, Lancaster, UK}
\affil[4]{IS-Instruments Ltd. 220 Vale Road, Tonbridge, Kent, UK}
\affil[5]{University of Kent, Canterbury, Kent, UK}
\date{}                     

\setcounter{Maxaffil}{0}
\renewcommand\Affilfont{\itshape\small}

\maketitle

\begin{abstract}
    Machine learning methods have found many applications in Raman spectroscopy, especially for the identification of chemical species. 
However, almost all of these methods require non-trivial preprocessing such as baseline correction and/or PCA as an essential step. Here we describe our unified solution for the identification of chemical species in which a convolutional neural network is trained to automatically identify substances according to their Raman spectrum without the need of ad-hoc preprocessing steps.
We evaluated our approach using the RRUFF spectral database, comprising mineral sample data. Superior classification performance is demonstrated compared with other frequently used machine learning algorithms including the popular support vector machine. 
\end{abstract}


\section{Introduction}


Raman spectroscopy is a ubiquitous method for characterisation of substances in a wide range of settings including industrial process control, planetary exploration, homeland security, life sciences, geological field expeditions and laboratory materials research.
In all of these environments there is a requirement to identify substances from their Raman spectrum at high rates and often in high volumes. Whilst machine classification has been demonstrated to be an essential approach to achieve real time identification, it still requires preprocessing of the data. This is true regardless of whether peak detection or multivariate methods, operating on whole spectra, are used as input.
A standard pipeline for a machine classification system based on Raman spectroscopy includes preprocessing in the following order: cosmic ray removal, smoothing and baseline correction. Additionally, the dimensionality of the data is often reduced using principal components analysis (PCA) prior to the classification step.  To the best of our knowledge, there is no existing work describing machine classification systems that can directly cope with raw spectra such as those affected significantly by baseline distorsion.

In this work we focus on multivariate methods, and introduce the application of convolutional neural networks (CNNs) in the context of Raman spectroscopy. Unlike the current Raman analysis pipelines, CNN combines preprocessing, feature extraction and classification in a single architecture which can be trained end-to-end with no manual tuning. We show that CNN not only greatly simplifies the development of a machine classification system for Raman spectroscopy, but also achieves significantly higher accuracy. In particular, we show that CNN trained on raw spectra significantly outperformed other machine learning methods such as support vector machine (SVM) with baseline corrected spectra.
Our method is extremely fast with a processing rate of one sample per millisecond \footnote{Software processing time only. Not including acquisition of Raman signal from spectrometer.}.


The baseline component of a Raman spectrum is caused primarily by fluorescence, can be more intense than the actual Raman scatter by several orders of magnitude, and adversely affects the performance of machine learning systems. Despite considerable effort in this area, baseline correction remains a challenging problem, especially for a fully automatic system\cite{Lieber03}.

A variety of methods for automatic baseline correction have been used such as polynomial baseline modelling\cite{Lieber03}, simulation-based methods\cite{KNEEN1996209, SiegBook2005}, penalized least squares\cite{zhang2010baseline, baek2015baseline}. Lieber et al.\cite{Lieber03} proposed a modified least-squares polynomial curve fitting for fluorescence substraction which was shown to be effective. Eilers et al.\cite{Paul2005} proposed a method called \textit{asymmetric least square smoothing}. One first smooths a signal by a Whittaker smoother to get an initial baseline estimation, and then applies asymmetric least square fitting where positive deviations with respect to baseline estimate are weighted (much) less than negative ones. This has been shown to be a useful method, and in principle can be used for automatic baseline correction, although it may occasionally  require human input.
Kneen et al.\cite{KNEEN1996209} proposed a method called \textit{rolling ball}. In this method one imagines a ball with tunable radius rolling over/under the signal. The trace of its lowest/highest point is regarded as an estimated baseline. A similar methods is  \textit{rubber band}\cite{SiegBook2005} where one simulates a rubber band to find the convex hull of the signal which can then be used as a baseline estimation.
Zhang et al.\cite{zhang2010baseline} presented a variant of penalized least squares, called \textit{adaptive iteratively reweighted Penalized Least Squares (airPLS)} algorithm. It iteratively adapts weights controlling the residual between the estimated baseline and the original signal.
A detailed review and comparison of baseline correction methods can be found in  Schulze et al.\cite{schulze2005investigation}.

Classification rates have been compared for various machine learning algorithms using Raman data. The method that is frequently reported to outperform other algorithms is support vector machines (SVM) \cite{vapnik1995nature}. An SVM is trained by searching for a hyperplane that optimally separates labelled training data with maximal margin between the training samples and the hyperplane. Binary (two class) and small scale problems in Raman spectroscopy have been previously addressed using this method. A large proportion of these relate to applications in the health sciences, use a non-linear SVM with a radial basis function kernel, and an initial principal component analysis (PCA) data reduction step. In this context SVM was shown to: outperform linear discriminant analysis (LDA) and partial least squares discriminant analysis (PLS-LDA) in breast cancer diagnosis \cite{Sattlecker2010}, successfully sort unlabelled biological samples into one of three classes (normal, hyperplastic polyps or adeno-carcinomas) \cite{Widjaja2008} and discriminate between three species of bacteria using a small number of training and test examples \cite{Kyriakides2009}. Although multiclass classification is possible using SVM, in practice training an non-linear SVM is infeasible for large scale problems involving thousands of classes. Random forests (RF) \cite{ho1998random} represent a viable alternative to SVM for high dimensional data with a large number of training examples. RF is an ensemble learning method based on multiple decision trees that avoids over fitting the model to the training set. This method generated a lot of attention in the machine learning community in last decade prior the widespread popularity of CNN. However when compared with PCA-LDA and RBF SVM on Raman microspectroscopy data \cite{Maguire2015} it performed poorly. The method previously applied to spectral classification problems that is closest to our own approach is fully connected artificial neural networks (ANN).
Unlike CNN, ANN is a shallow architectures which does not have enough capacity to solve large scale  problems. Maquel et al. \cite{Maquelin2003} determined the major groupings for their data prior to a multilayered ANN analysis. Their study concluded that vibrational spectroscopic techniques are well suited to automatic classification and can therefore be used by nonexperts and at low cost.

\newcolumntype{b}{X}
\newcolumntype{t}{>{\hsize=.5\hsize}X}
\newcolumntype{a}{>{\hsize=1.2\hsize}X}
\renewcommand{\arraystretch}{1.5}
\begin{wraptable}{r}{0.6\columnwidth}
\centering
\footnotesize
\caption{Summary of problems of ours and those in existing works}
\label{Tab:CompareMlProblems}
\begin{threeparttable}
\begin{tabularx}{0.6\columnwidth}{bccc}
\toprule
\textbf{Problems} & \#Classes & \#Spectra & Baseline Removal  \\
\midrule
Sattlecker et al.\cite{Sattlecker2010}&  2 & 1905 &  N/A$^\clubsuit$ \\
Kwiatkowski et al.\cite{kwiatkowski2010algorithms}& 10 & N/A & Yes \\
Carey et al.\cite{carey2015machine} &  1215 & 3950 & Yes \\
Ours \#1 & 1671 & 5168 &  Yes  \\
Ours \#2 & 512 & 1676  & No  \\
\bottomrule
\end{tabularx}
\begin{tablenotes}
	\vspace{0.25cm}
    \item[1] $\clubsuit$ Note that in this work special filtering methods were developed to discard spectra of bad quality which account for 80\% of total amount.
  \end{tablenotes}
\end{threeparttable}
\end{wraptable}

A drawback associated with the methods previously used is that they require feature engineering (or preprocessing) and don't necessarily scale easily to problems involving a large number of classes. Motivated by the recent and widespread success of CNNs in large scale image classification problems we developed our network architecture for the classification of 1D spectral data. A suitable dataset to test the efficacy of the CNN is the RRUFF mineral dataset. Previous work \cite{Ishikawa2013AMC,carey2015machine} has focused on identifying mineral species contained in this dataset using nearest neighbor methods with different similarity metrics such as cosine similarity and correlation (also used in commercial softwares such as CrystalSleuth). Carey et al.\cite{carey2015machine} achieved a species classification accuracy on a subset of the RRUFF database \cite{RRUFFdataset} of 84.8\% using a weighted neigbour (WN) classifier. Square root squashing, maximum intensity normalisation, and sigmoid transformations were applied to the data prior to classification. Accuracy was determined using cross validation with semi-randomised splits over a number of trials. 
The WN classifier compared favourably with the $k=1$  nearest neighbour (82.1\% accuracy) on which the CrystalSleuth matching software is believed to be based. 
In Table~\ref{Tab:CompareMlProblems} we summarise the sample data used in our own work and in some previous Raman based spectral classification studies.

\section{Materials and Methods}\label{MaterialsAndMethods}

\begin{figure*}
	\centering
	\includegraphics[width=\columnwidth, angle=0]{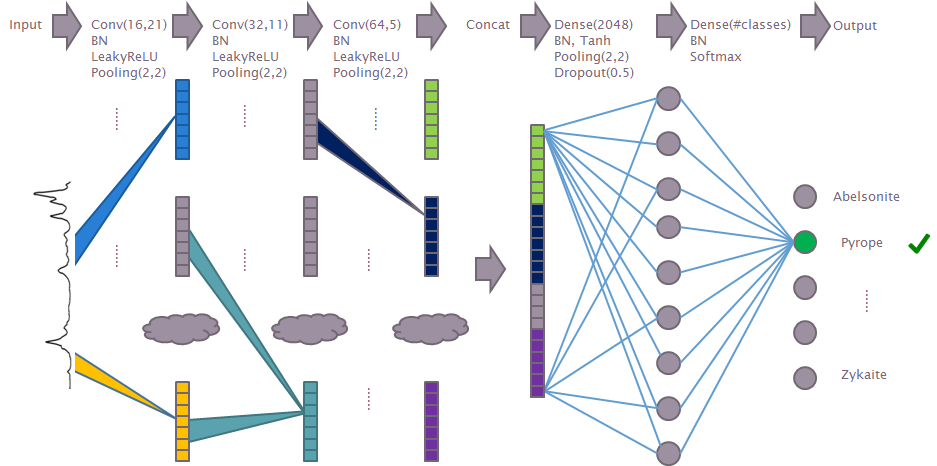}	
	\caption{Diagram of the proposed CNN for spectrum recognition. It consists of a number of convolutional layers for feature extraction and two fully-connected layer for classification.}
	\label{Fig:smallcnn}
\end{figure*}

CNNs have become the predominant tool in a number of research areas - especially in computer vision and text analysis. An extension of the artificial neural network concept \cite{Hubel1968}, CNNs are nonlinear classifiers that can identify unseen examples without the need for feature engineering.
They are computational models\cite{lecun1998gradient} inspired by the complex arrangement of cells in the mammalian visual cortex. These cells are stimulated by small regions of the visual field, act as local filters, and encode spatially localised regions of natural signals or images.

CNNs are designed to extract features from an input signal with different levels of abstraction. A typical CNN includes convolutional layers, which learn filter maps for different types of patterns in the input, and pooling operators which extract the most prominent structures. The combination of convolutional and pooling layers extracts features (or patterns) hierarchically. Convolutional layers share weights which allow computations to be saved and also make the classifier invariant to spatial translation. The fully connected layers (that follow the convolutional and pooling layers) and the softmax output layer can be viewed as a classifier which operates on the features (of the Raman spectra data), extracted using the convolutional and pooling layers. Since all layers are trained together, CNNs integrate feature extraction with classification. Features determined by network training are optimal in the sense of  the performance of the classifier.  Such end-to-end trainable systems offer a much better alternative to a pipeline in which each part is trained independently or crafted manually.


In this work, we evaluated the application of a number of prominent CNN architectures including LeNets\cite{Lecun98gradientbasedlearning}, Inception\cite{InceptionV1} and Residual Nets\cite{he2016deep} to Raman spectral data. All three showed comparable classification results even though the latter two  have been considered superior to LeNet in computer vision applications.  We  adopted a variant of LeNet, comprising pyramid-shaped convolutional layers for feature extraction and two fully-connected layers for classification. A graphical illustration of the network is shown in Figure \ref{Fig:smallcnn}.

\subsection{CNN for Raman Spectral Data Classification}
The input to the CNN in application to Raman spectrum classification is one dimensional and it contains the entire spectrum (intensity fully sampled at regularly spaced wavenumbers).  Hence we trained one-dimensional convolutional kernels in our CNN.  For the convolutional layers, we used LeakyReLU\cite{maas2013rectifier} nonlinearity, defined as
\begin{equation}
    f(x)=
    \begin{cases}
      x, & \text{if}\ x>0 \\
      ax, & \text{otherwise}
    \end{cases}
  \end{equation}
Formally, a convolutional layer can be expressed as follows:
\begin{equation}
    y^j=f\left(b^j+\sum_i{k^{ij}*x^i}\right)
\end{equation}
where $x^i$ and $y^i$ are the $i$-th input map and the $j$-th output map, respectively. $k^{ij}$ is a convolutional kernel between the maps $i$ and $j$, $*$ denotes convolution, and $b^j$ is the bias parameter of the $j$-th map.

The convolutional layer is followed by a max-pooling layer, in which each neuron in the output map $y^i$ pools over an $s \times s$ non-overlapping region in the input map $x^i$. Formally,
\begin{equation}
    y^i_{j}=\max_{0\leq m<s}\{x^i_{j\cdot s+m}\}
\end{equation}

The upper layers of the CNN are fully connected and followed by the softmax with the number of outputs equal to the number of classes considered. We used $tanh$ as non-linearity in the fully connected layers. The softmax operates as a squashing function that re-normalizes a K-dimensional input vector $z$ of real values to  real values in the range $[0, 1]$ that sum to 1, specifically,
\begin{align}
\sigma(z)_j=\frac{e^{z_j}}{\sum_{k=1}^K e^{z_k}}\;\;\;\textnormal{for}\;\;j=1,...,K.
\end{align}

To avoid overfitting the model to the data, we applied batch normalization~\cite{2015arXiv150203167I} after each layer and dropout~\cite{Srivastava:2014} after the first fully connected layer. Further details of the architecture are shown in Figure~\ref{Fig:smallcnn}.

\subsection{CNN Training}
Since the classes in the our experiments have very different numbers of examples, we used the following weighted loss to train the CNN:
\begin{align}
		\mathcal{L}(\textbf{w},x_n,y_n) = -  \frac{1}{N} \sum_{n=1}^{N} \alpha_{n} \sum_{k=1}^{K} t_{kn} \ln y_{kn}
\end{align}
where $x_n$ is a training spectrum, $t_{n}$ is the true label of the n$^{th}$ sample in the format of one-hot encoding, $y_{n}$ is the network prediction for the n$^{th}$ sample, $\alpha_{n} \propto \frac{1}{\#C}$ and $\#C$ is the number of samples in the class $C$ that $x_{n}$ belongs to. $N$ is the total number of samples and $K$ is the number of the classes.

CNN is a data hungry model. To reduce the data volume requirements we use augmentation which is a very common approach for increasing the size of the training sets for CNN training. Here, we propose the following data augmentation procedure: (1) We shifted each spectrum left or right a few wavenumbers randomly. (2) We added a random noise, proportional to the magnitude at each wave number. (3) For the substances which had more than one spectra, we took linear combinations of all spectra belonging to the same substance as augmented data. The coefficients in the linear combination were chosen at random.

The training of the CNN was performed using Adam algorithm~\cite{kingma2014adam}, which is a variant of stochastic gradient descent, for 50 epochs with learning rate equal to 1e-3, $\beta_{1} = 0.9$, $\beta_{2}= 0.999$, and $\epsilon$=1e-8. The layers were initialised from a Gaussian distribution with a zero mean and variance equal to 0.05. We applied early stopping to prevent overfitting.  Training was performed on a single NVIDIA GTX-1080 GPU. The training time was around seven hours. While for inference, it took less than one millisecond to process a spectrum.





\subsection{Evaluation Protocol}\label{Sec_Eva_Protocol}
We tested the proposed CNN method for mineral species recognition on the largest publicly available mineral database RRUFF\cite{RRUFFdataset} and compared it  with a number of alternative, well known, machine learning methods.
As there are usually only a handful of spectra available for each mineral, we use a leave-one-out scheme to split a dataset into training and test sets. To be specific, for minerals which have more than one spectra, we randomly select a spectrum for testing and use the rest for training. We compared our method to cosine similarity~\cite{carey2015machine}/correction~\cite{kwiatkowski2010algorithms} (which has been used in commercial software such as CrystalSleuth and Spectral-ID), and to other methods that have been shown to be successful in classification tasks including applications based on Raman: nearest neighbor, gradient boosting machine, random forest, and support vector machine\cite{bishop2006pattern}.

The proposed CNN was implemented using Keras\cite{chollet2015keras} and Tensorflow\cite{tensorflow2015-whitepaper}. The gradient boosting machine method was implemented based on lightGBM released by Microsoft. All other methods were implemented using on Scikit-learn\cite{sklearn_api}.

\section{Results and Discussion}\label{ExpResults}

\subsection{Classifying baseline-corrected spectra}\label{Sec_RRUFF}

\newcolumntype{b}{X}
\newcolumntype{s}{>{\hsize=.6\hsize}X}
\renewcommand{\arraystretch}{1.5}

\begin{table*}
\caption{Test accuracy of the compared machine learning methods on the baseline corrected dataset}
\label{Tab:SummaryRruffExUo}
\footnotesize
\centering
\begin{tabularx}{\textwidth}{bssssssc}
\hline
\textbf{Methods} & KNN(k=1) & Gradient Boosting & Random Forest$\dagger$ & SVM(linear) & SVM(rbf) & Correlation & CNN$\dagger$ \\ \midrule
\textbf{Top-1 Accuracy} & 0.779{\footnotesize $\pm$0.011} & 0.617{\footnotesize $\pm$0.008} & 0.645{\footnotesize $\pm$0.007}
& 0.819{\footnotesize $\pm$0.004} &0.746{\footnotesize $\pm$0.003} & 0.717{\footnotesize $\pm$0.006} &\textbf{0.884{\footnotesize$\pm$0.005}}\\
\textbf{Top-3 Accuracy} & 0.780{\footnotesize $\pm$0.011} & 0.763{\footnotesize $\pm$0.011}  &0.753{\footnotesize $\pm$0.010}
&0.903{\footnotesize $\pm$0.006} &0.864{\footnotesize $\pm$0.006}& 0.829{\footnotesize $\pm$0.005} & \textbf{0.953{\footnotesize$\pm$0.002}}\\
\textbf{Top-5 Accuracy} & 0.780{\footnotesize $\pm$0.011} & 0.812{\footnotesize $\pm$0.010}  & 0.789{\footnotesize $\pm$0.009}
&0.920{\footnotesize $\pm$0.003}&0.890{\footnotesize $\pm$0.007}& 0.857{\footnotesize $\pm$0.005} &\textbf{0.963{\footnotesize $\pm$0.002}}\\ \hline
\end{tabularx}
\end{table*}

\begin{figure}[!htp]
    \subfigure[Spectra of \textit{Actinolite}\cite{RRUFFdataset}.]{
	\includegraphics[width=0.475\columnwidth]{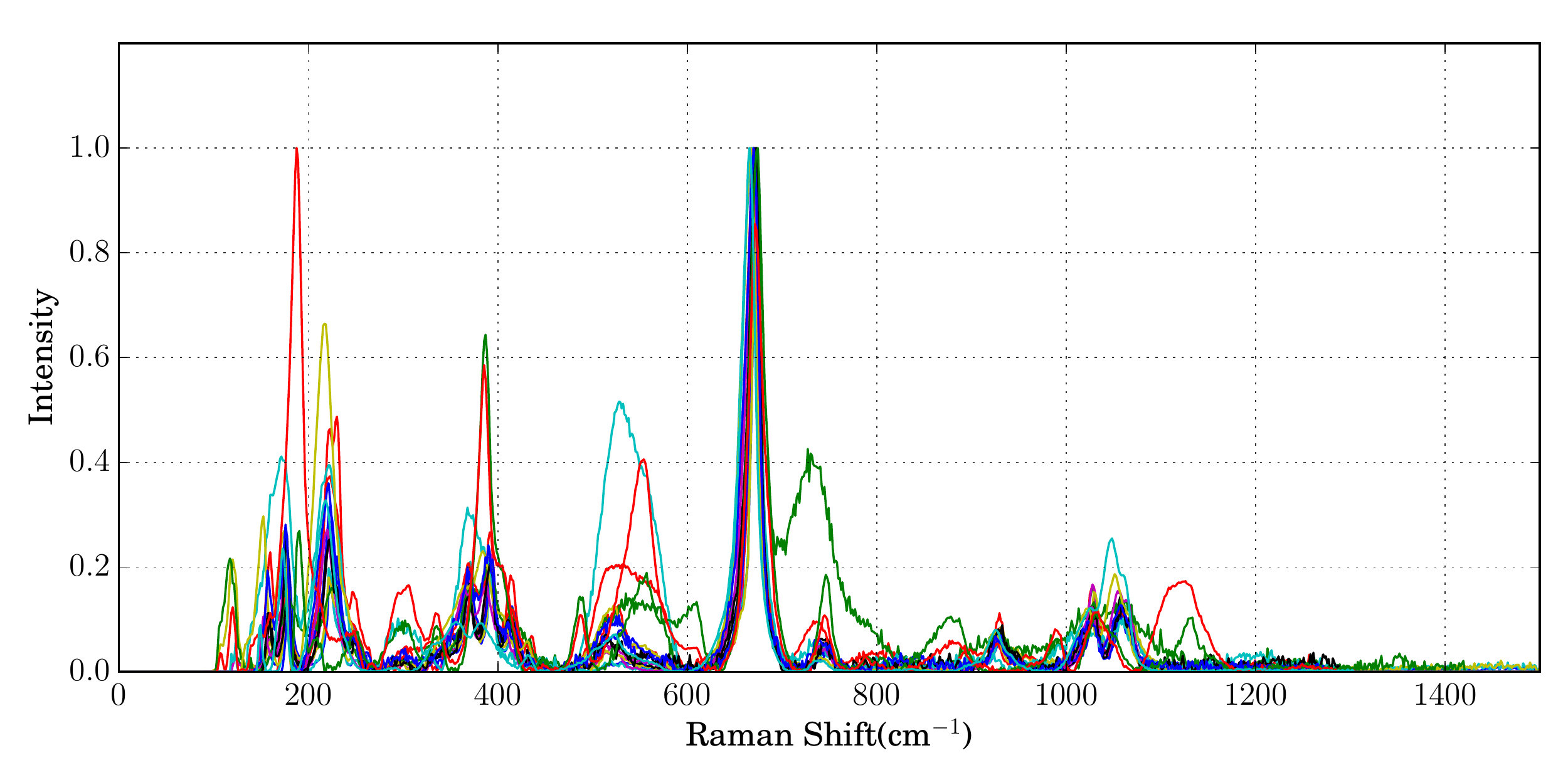}	
    }
    \subfigure[Number of spectra per mineral of the whole dataset.]{
    \includegraphics[width=0.475\columnwidth]{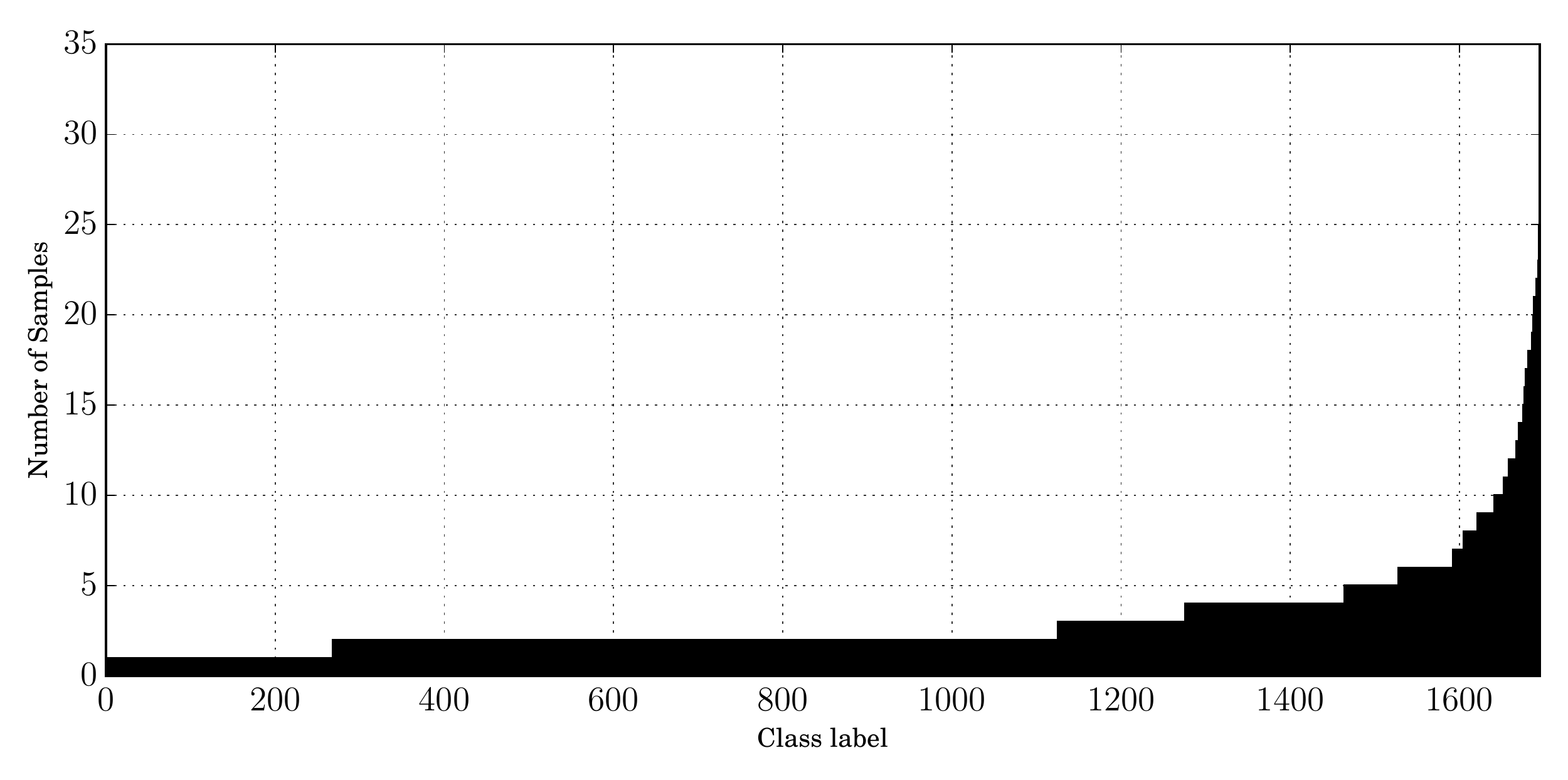}
    }
	\caption{(a) Spectra of an example mineral species (\textit{Actinolite}) indicating the within class spectrum variation and (b) a frequency plot showing the imbalance regarding spectra per species.}
	\label{Fig:mineral1700}
\end{figure}

We first evaluated our CNN method on a processed mineral dataset from the RRUFF database. These spectra have been baseline corrected and cosmic rays have also been removed. The dataset contains 1671 different kinds of minerals, 5168 spectra in total. Spectra for the mineral \textit{Actinolite} is shown in Figure~\ref{Fig:mineral1700}(a), illustrating the typical within-class variance.
The number of spectra per mineral ranges from 1 to 40. The distribution of sample numbers per a mineral species is shown in Figure \ref{Fig:mineral1700}(b).
We followed the protocol as described in Section \ref{Sec_Eva_Protocol} to generate training and test sets randomly using the leave-one-out scheme.

In a large scale classification, some classes could be quite similar and differentiating between them could be very difficult or even impossible. Hence, it is common to report top-1 and top-k accuracy. In the former, the class that the classifier assigns the highest probability to is compared to the true label. The latter reports whether the true label appears among the k classes with the highest probability (assigned by the classifier).

We report in Table \ref{Tab:SummaryRruffExUo}, the top 1, 3 and 5 accuracies of the compared methods, averaged over 50 independent runs. One can see that CNN significantly outperformed all other methods and achieved top-1 accuracy of 88.4\% and top-3 accuracy of 96.3\%. 

\begin{figure*}[!htp]
	\centering
    \subfigure[Succeeded, top-1 hit]{
	\includegraphics[width=0.475\columnwidth]		
    	{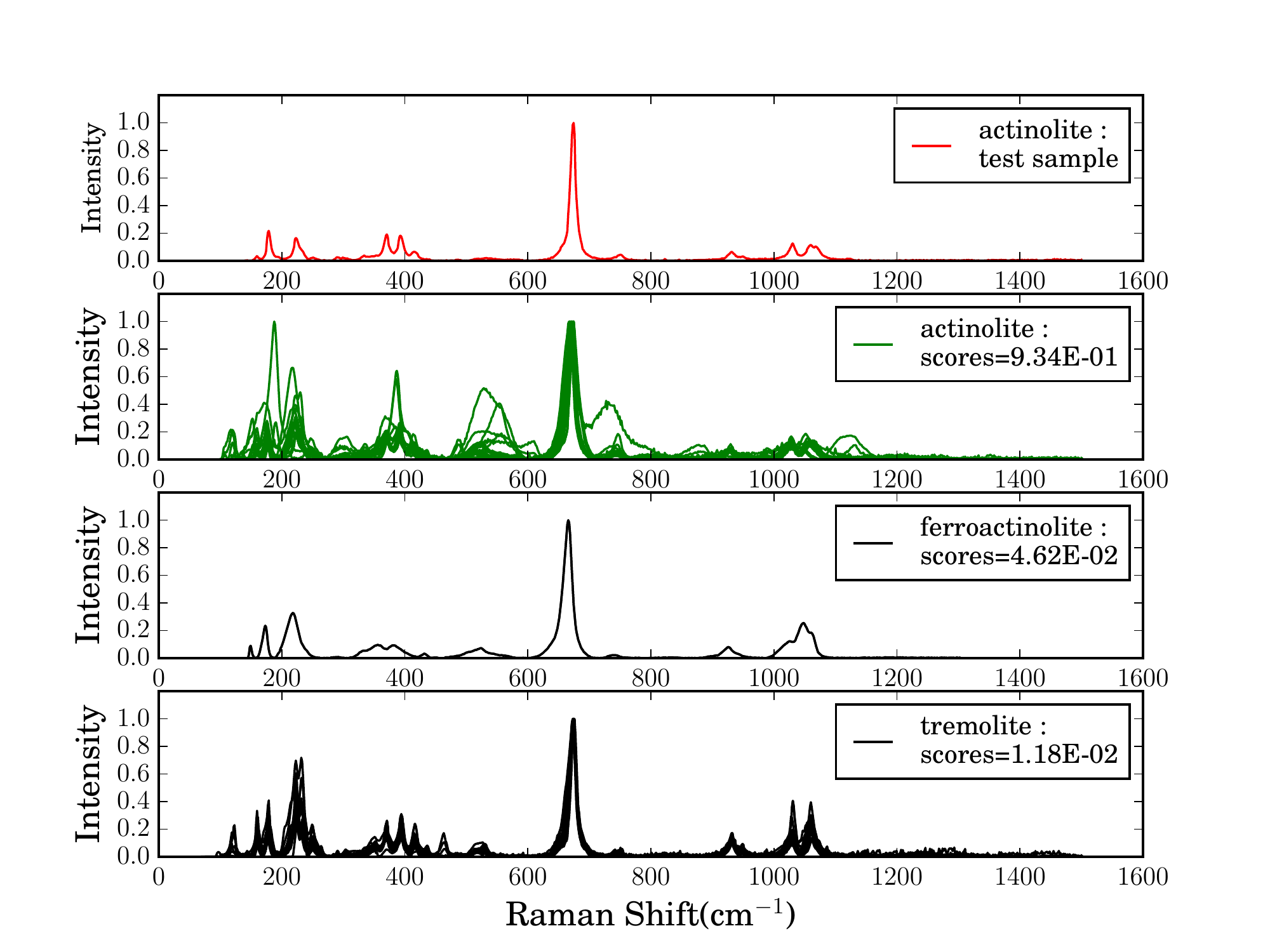}	
    }
    \subfigure[Succeeded, top-3 hit]{
    	\includegraphics[width=0.475\columnwidth]
    	{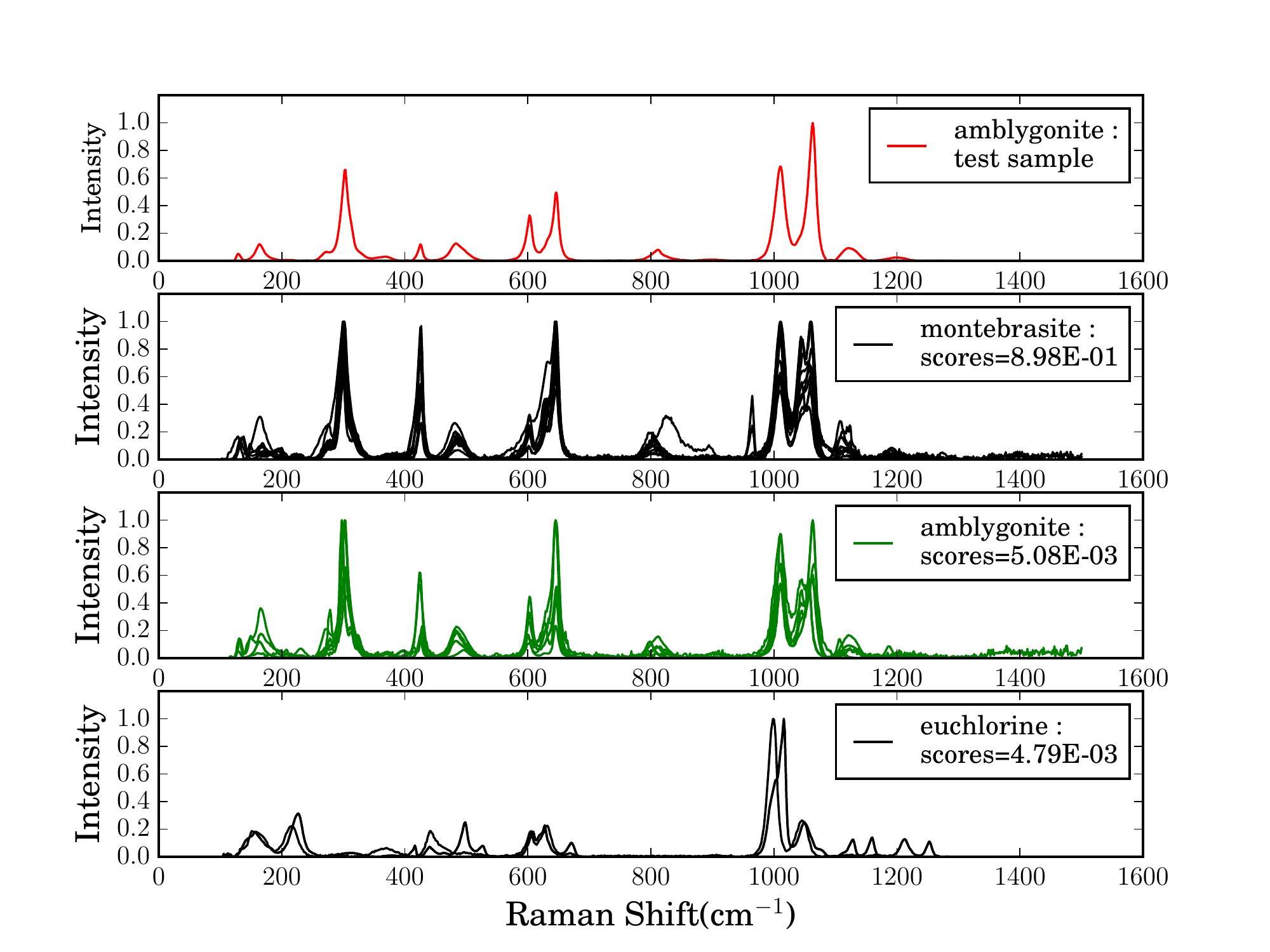}	
    }
    \subfigure[Failed, but similar]{
    	\includegraphics[width=0.475\columnwidth]
    	{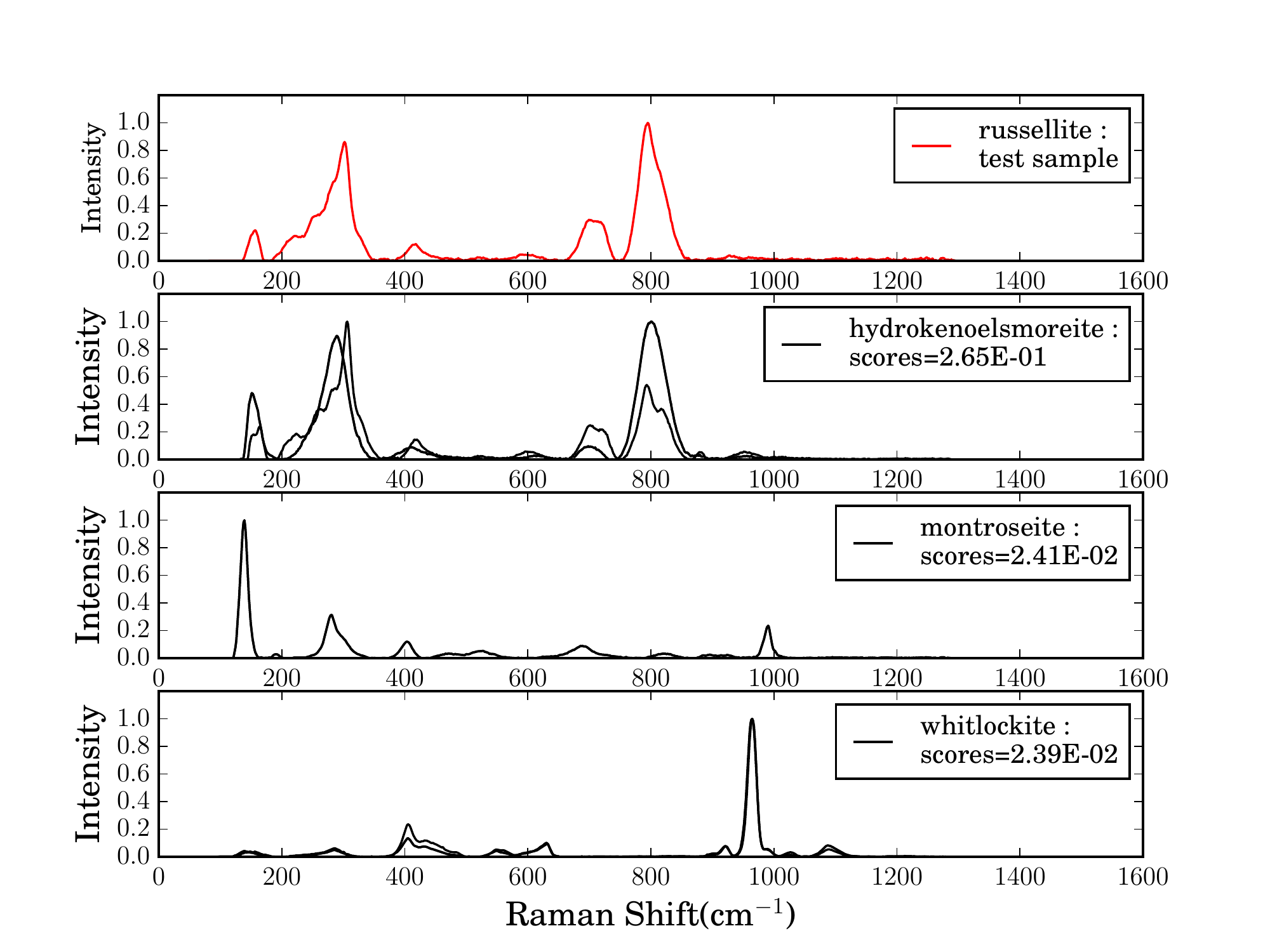}	
    }
    \subfigure[Failed, partially matching]{
    	\includegraphics[width=0.475\columnwidth]
    	{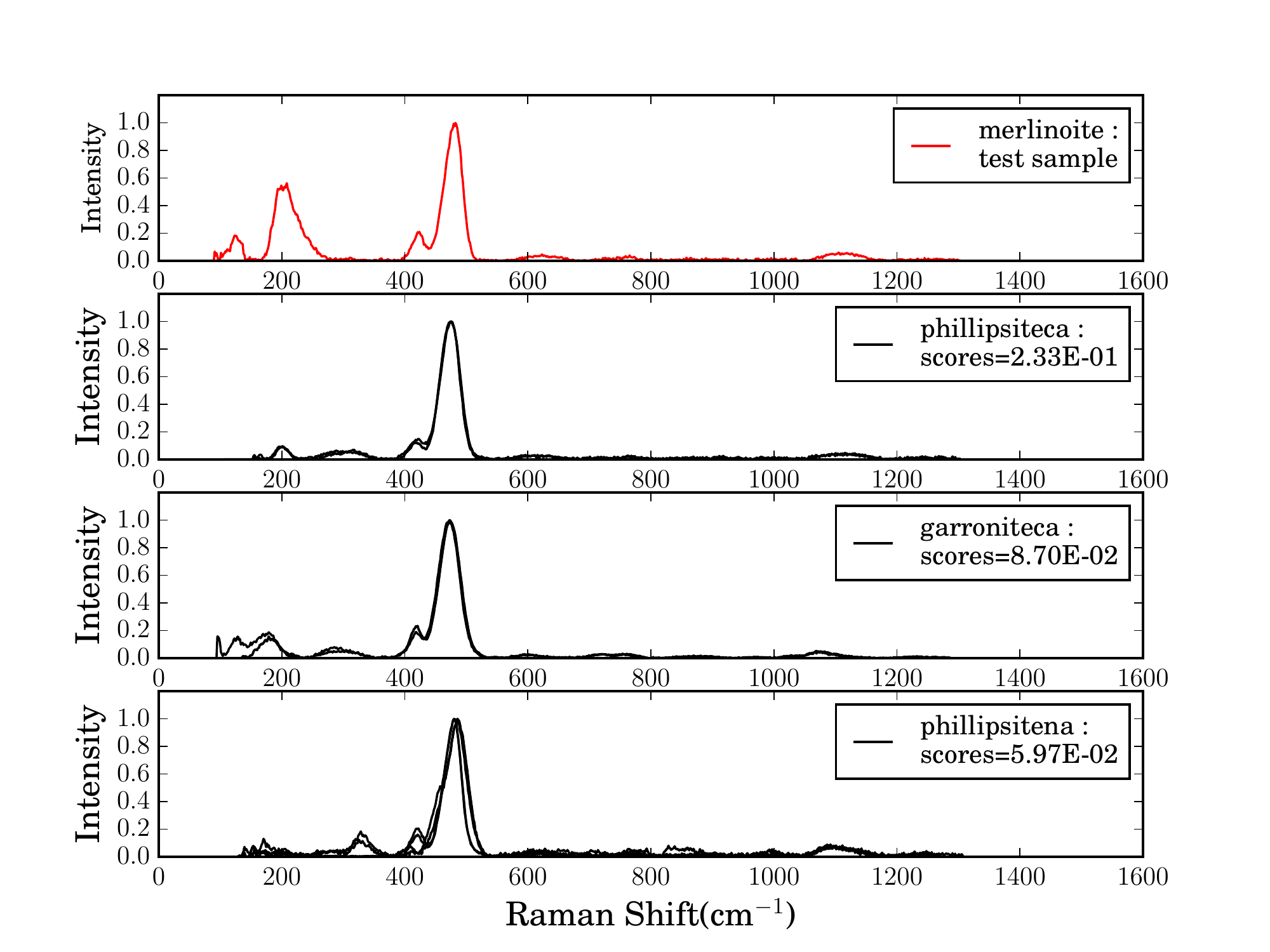}	
    }
	\caption{Examples of successful and unsuccessful mineral species classifications. In each plot, the top spectrum which is marked in red is a test sample. The three spectra below were the top-3 predictions given by the CNN among which the correct one was highlighted in green. The prediction scores were also shown in each plot which reflect the confidence level of predictions.}
	\label{Fig:Top3hits}
\end{figure*}


\renewcommand{\arraystretch}{1.5}
\begin{table*}
\footnotesize
\centering
\caption{Test accuracy of the compared machine learning methods on raw dataset with or without baseline correction methods}
\label{Tab:SummaryRRUFFCMP}
\begin{tabularx}{\textwidth}{bssssssc}
\toprule
\textbf{Methods} & KNN(k=1) & Gradient Boosting & Random Forest$\dagger$ & SVM(linear) & SVM(rbf) & Correlation & CNN$\dagger$ \\ \midrule
\textbf{Raw} & 0.429{\footnotesize $\pm$0.011} & 0.373{\footnotesize $\pm$0.019} & 0.394{\footnotesize $\pm$0.016} & 0.522{\footnotesize $\pm$0.011} & 0.434{\footnotesize $\pm$0.012} & 0.310{\footnotesize $\pm$0.007} &  \textbf{0.933{\footnotesize $\pm$0.007}}\\

\textbf{Asym LS} & 0.817{\footnotesize $\pm$0.010} & 0.773{\footnotesize $\pm$0.009} & 0.731{\footnotesize $\pm$0.019} & 0.821{\footnotesize $\pm$0.012} & 0.629{\footnotesize $\pm$0.016} & 0.777{\footnotesize $\pm$0.013} &  \textbf{0.927{\footnotesize $\pm$0.008}}\\

\textbf{Modified Poly} & 0.778{\footnotesize $\pm$0.007} & 0.740{\footnotesize $\pm$0.016} & 0.650{\footnotesize $\pm$0.016} & 0.785{\footnotesize $\pm$0.014} & 0.629{\footnotesize $\pm$0.016} & 0.734{\footnotesize $\pm$0.013} &  \textbf{0.920{\footnotesize $\pm$0.008}}\\

\textbf{Rolling Ball} & 0.775{\footnotesize $\pm$0.009} & 0.737{\footnotesize $\pm$0.008}  & 0.689{\footnotesize $\pm$0.018} & 0.795{\footnotesize $\pm$0.011} & 0.624{\footnotesize $\pm$0.013} & 0.730{\footnotesize $\pm$0.010} &  \textbf{0.918{\footnotesize $\pm$0.008}}\\

\textbf{Rubber Band} & 0.825{\footnotesize $\pm$0.007} & 0.792{\footnotesize $\pm$0.015} & 0.741{\footnotesize $\pm$0.009} & 0.806{\footnotesize $\pm$0.015} & 0.620{\footnotesize $\pm$0.010} & 0.789{\footnotesize $\pm$0.010} &  \textbf{0.911{\footnotesize $\pm$0.008}}\\

\textbf{IRLS} & 0.772{\footnotesize $\pm$0.010} & 0.710{\footnotesize $\pm$0.008} & 0.675{\footnotesize $\pm$0.007} & 0.781{\footnotesize $\pm$0.011} & 0.614{\footnotesize $\pm$0.010} & 0.711{\footnotesize $\pm$0.011} &  \textbf{0.911{\footnotesize $\pm$0.008}}\\

\textbf{Robust LR} & 0.741{\footnotesize $\pm$0.009} & 0.694{\footnotesize $\pm$0.008} & 0.667{\footnotesize $\pm$0.012} & 0.759{\footnotesize $\pm$0.013} & 0.600{\footnotesize $\pm$0.013} & 0.696{\footnotesize $\pm$0.011} &  \textbf{0.909{\footnotesize $\pm$0.007}}\\
\bottomrule
\end{tabularx}
\end{table*}

To understand the trained model of CNN better, we also closely examined typical predictions, especially where these did not agree with the correct labelling. In Figure \ref{Fig:Top3hits} the top spectrum in each set is the test sample (shown in red) which is followed by the top-3 predictions given by the CNN. The correct prediction is highlighted in green. We also show scores in each plot which reflect the confidence level of predictions. Figure \ref{Fig:Top3hits}(a) shows the examples where the CNN made the correct prediction. Figure \ref{Fig:Top3hits}(b) shows the examples in which the correct prediction is scored second. In Figure \ref{Fig:Top3hits}(c), the top-3 predictions do not include the correct label.

As shown in Figure \ref{Fig:Top3hits}(a), the CNN successfully predicted the correct mineral, \textit{actinolite}, and also ranked \textit{Ferroactinolite} and \textit{Tremolite} as the second and third probable candidates. 
In fact, all these three minerals are members of the same mineral group. This is not uncommon. For instance, in Figure \ref{Fig:Top3hits}(b), the most probable mineral \textit{Montebrasite} (as predicted by the CNN) belongs to the same group as the correct one, \textit{Amblygonite}, and they share similar spectral structure.

If we examine the peak similarity, for instance in Figure \ref{Fig:Top3hits}(c), the peak locations of the top-1 prediction, \textit{Hydrokenoelsmoreite}, are almost identical to those of the test sample \textit{Russellite}. 
In Figure \ref{Fig:Top3hits}(d), only the main peaks were matched correctly.
These plots demonstrate that the CNN was capable of matching the peaks characteristic of a particular species even when the prediction did not agree with the correct label.


\begin{figure}[!htp]
    \subfigure[Ten raw spectra where baselines can be clearly observed.]{
	\includegraphics[width=0.475\columnwidth]{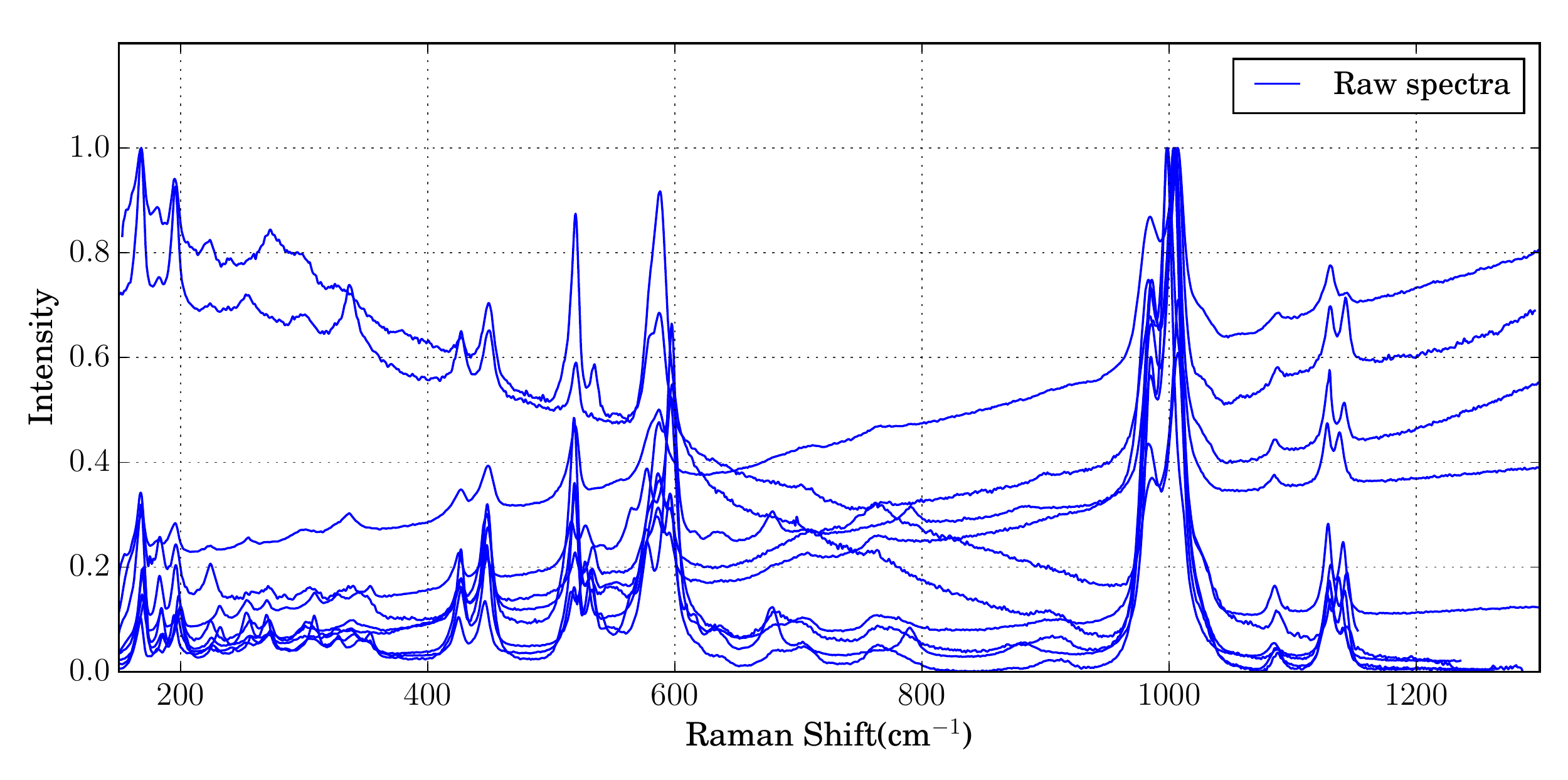}
    }
    \subfigure[Baseline corrected by asymmetric least squares]{
    \includegraphics[width=0.475\columnwidth]{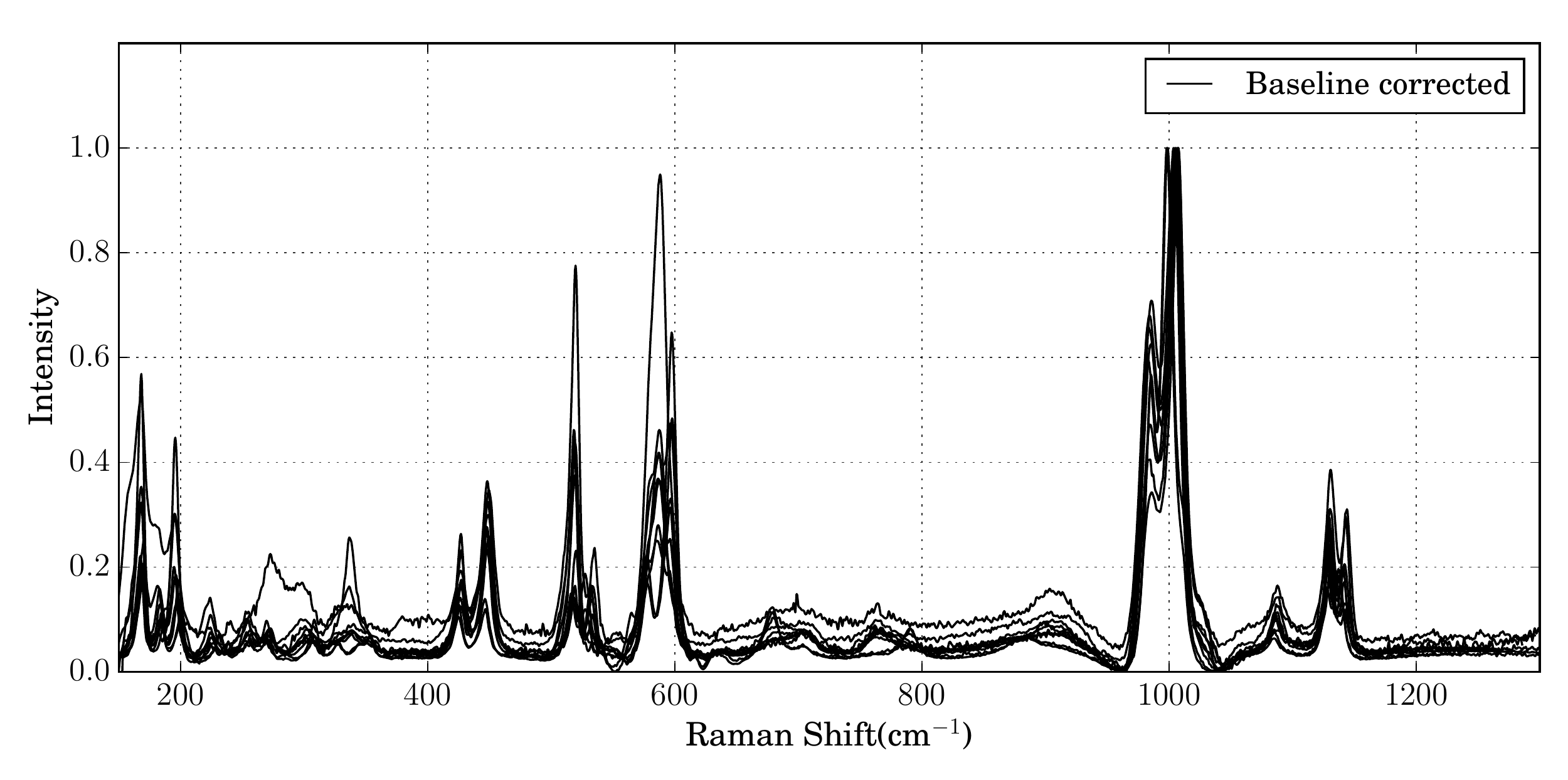}
    }
	\caption{Spectra of a mineral, \textit{hydroxylherderite}, from RRUFF raw database and corresponding baseline corrected ones by asymmetric least squares.}
	\label{Fig:RawBaseline}
\end{figure}

\subsection{Unified Raman Analysis using CNN} 
The results in Section \ref{Sec_RRUFF} have shown that CNN was able to achieve significantly better accuracy compared to other conventional machine learning methods on the \textit{baseline-corrected} spectra. Recall that conventional machine learning methods such as SVM and Random Forest are not capable of handling Raman signals which are not properly baseline corrected, and therefore require explicit baseline correction in their processing pipelines. However, robust baseline correction is a challenging problem, especially for a fully automatic system\cite{Lieber03}.
On the other hand, in a variety of applications, CNN has been shown to be very successful as an end-to-end learning tool since it can process the data and learn features automatically, avoiding hand crafted feature engineering\cite{krizhevsky2012imagenet}.
Therefore, we evaluated the proposed CNN and and the other classification methods using both raw and baseline corrected spectra. Specifically, we were interested in the performance of CNN on raw data compared to the previous state-of-the-art spectral classifaction methods for which baseline correction \emph{was included}.

For this set of experiments, we selected another dataset from the RRUFF database which contains raw (uncorrected) spectra for 512 minerals and six widely-used baseline correction methods:
\textit{modified polynomial fitting\cite{Lieber03}, rubber band\cite{SiegBook2005}, robust local regression estimation\cite{RUCKSTUHL2001179}, iterative restricted least squares, asymmetric least square smoothing\cite{Paul2005}, rolling ball\cite{KNEEN1996209}}.
We used implementations of these methods in the R packages \textit{baseline}\cite{R_baseline} and \textit{hyperSpec}\cite{R_hyperSpec}.
An example of raw spectra and corresponding baseline corrected ones by asymmetric least squares is shown in Figure \ref{Fig:RawBaseline}.
We followed the training and evaluation protocol as described in Section \ref{Sec_Eva_Protocol}. The results are reported in Table \ref{Tab:SummaryRRUFFCMP}.


For the conventional classification methods, used as a comparison in our work, PCA was adopted to reduce dimensionality and extract features, except for Random Forest where we found that PCA decreased the performance. This is indicated in the table by $\dagger$. The number of principal components were determined such that 99.9\% of total variance was retained.
One can see that CNN on the raw spectra achieved an accuracy of 93.3\% which is significantly better than the second best method, KNN with rubber band baseline correction, with accuracy 82.5\%.

There are a few remarks which are worth highlighting. Firstly, it is not a surprise that baseline correction greatly improved the performance of all the conventional methods by 20\% $\sim$ 40\%. On other hand, CNN's performance dropped by about 0.5\% $\sim$ 2.5\% when combined with baseline correction methods. This may indicate that CNN was able to learn more efficient way of handling the interference of the baselines and to retain more discriminant information than using an explicit baseline correction method. The advantage of CNNs in achieving high accuracy of classification while requiring minimal preprocessing of spectra opens new possibilities for developing highly accurate fully automatic spectrum recognition systems.

\section{Conclusion and Future Work}\label{Conclusion}
In this paper, we have presented a deep convolutional neural network solution for Raman spectrum classification which not only exhibits outstanding performance, but also avoids the need for spectrum preprocessing of any kind. Our method has been validated on a large scale mineral database and was shown to outperform other state-of-the-art machine learning methods by a large margin. Although we focused our study on Raman data we believe the method is also applicable to other spectroscopy and spectrometry methods. We speculate that this may be acheived very efficiently by exploiting basic similarities in the shape of spectra originating from different techniques and fine tuning our network to address new classification problems. This process is known as transfer learning and has been demomstrated previously in many object recognition applications.

\bibliography{spec_classification}
\bibliographystyle{ieeetr}

\end{document}